# Can LLMs capture stable human-generated sentence entropy measures?

Estrella Pivel-Villanueva 1,2, Elisabeth Frederike Sterner 1,2,3, Franziska Knolle* 2,3

1 Graduate School of Systemic Neurosciences, Ludwig Maximilian University, Munich, Germany

2 Department of Diagnostic and Interventional Neuroradiology, School of Medicine and Health, Technical University of Munich, Munich, Germany

3 School of Medicine and Health, TUM-NIC Neuroimaging Center, Technical University of Munich, Munich, Germany

*Corresponding author:

Franziska Knolle: franziska.knolle@tum.de

**Abstract (280)**

Predicting upcoming words is a core mechanism of human language comprehension and is commonly quantified using Shannon entropy derived from cloze completion tasks. Despite the widespread use of entropy as a measure of contextual constraint, there is currently no empirical consensus on how many human responses are required to obtain stable and unbiased entropy estimates at the word level. Moreover, large language models (LLMs) are increasingly used as substitutes for human norming data, yet their ability to reproduce stable human entropy remains unclear.

Here, we address both issues using two large publicly available cloze datasets in German [1] and English [2]. We implemented a bootstrap-based convergence analysis that tracks how entropy estimates stabilize as a function of sample size. Across both languages, more than 97% of sentences reached stable entropy estimates within the available sample sizes. 90% of sentences converged after 111 responses in German and 81 responses in English, while low-entropy sentences (<1) required as few as 20 responses and high-entropy sentences (>2.5) substantially more. These findings provide the first direct empirical validation for common norming practices and demonstrate that convergence critically depends on sentence predictability.

We then compared stable human entropy values with entropy estimates derived from several LLMs, including GPT-4o, using both logit-based probability extraction and sampling-based frequency estimation, GPT2-xl/german-GPT-2, RoBERTa Base/GottBERT, and LLaMA 2 7B Chat. GPT-4o showed the highest correspondence with human data, although alignment depended strongly on the extraction method and prompt design. Logit-based estimates minimized absolute error, whereas sampling-based estimates were better in capturing the dispersion of human variability.

Together, our results establish practical guidelines for human norming and show that while LLMs can approximate human entropy, they are not interchangeable with stable human-derived distributions.

1. Introduction

Predicting the next word while listening to someone speak or while reading a sentence is a central feature of human language comprehension. Linguistic predictability or contextual constraint can be quantified using cloze tasks, in which participants predict the final word of a sentence [3]. Measures derived from these tasks, including Shannon entropy and surprisal [4], predict key behavioural and neural indicators of language processing such as reading time, eye movements, and ERP components like the N400 [5–8]. These measures have also been used in computational modelling and cognitive neuroscience approaches to understand psychiatric disorders or trait symptoms [9–12].

With the rise of large language models (LLMs) like GPT or BERT, researchers have access to probability distributions over word completions generated by powerful computational systems, trained on endless written text. This has prompted interest in replacing costly human data collection with automated predictions. However, studies that compared the two on different levels revealed consistent misalignments between human-based and LLM-based cloze responses [8,13–17]. Interestingly, LLMs seem to systematically assign lower probability to the most common human completions and greater probability to rare or semantically deviant alternatives [18,19]. Attempts to overcome these difficulties, through for example Cloze Distillation [20] or entropy-informed fine-tuning [21,22], offer partial improvements but fail to fully replicate human distributions.

There are different reasons for this persistent incompatibility – the most prominent may lie directly in the two systems that are being compared. In humans, semantic spaces emerge from the integration of language with sensory, motor, and emotional experiences, so that a concept like *apple* is linked to how it looks, tastes, and is used in daily life, and these associations vary across individuals and cultures. In contrast, the semantic space of a LLM is constructed solely from patterns of word co-occurrence in text. Thus, a word like *apple* is represented only through its statistical relationships to surrounding words, producing a coherent but disembodied and homogenised representation which lack the impact of personal experiences and variability. Furthermore, while human semantic spaces are large but constrained by cognitive and perceptual capacities, LLM semantic spaces are extraordinarily high-dimensional, encompassing millions of words and contexts learned from massive text corpora. Given these inherent differences it is questionable whether a LLM could ever fully replicate a human semantic space. Wang and colleagues [23] argue that lower interconnectivity and association organisation of semantic networks of current LLMs make them inferior to human semantic networks and limit their creativity and reasoning. Similarly, a study by Xiao and colleagues [24] reveals that while the semantic relatedness of human mental lexicons is represented by LLMs, LLMs show less diversity and greater clustering. In their word association task, they also showed that LLMs failed in representing high frequency associations [24]. Furthermore, in contrast to humans, who show highly personalised semantic styles, LLMs are "straightforward, factual and strictly-framed" [25] which might contribute to the differences and poorer performance in capturing associations in general or those to reflect specific demographic groups [24]. Complementing this view, recent work by Peng et al. [26] examined whether LLMs can predict human semantic feature norms rather than next-word probabilities across English and Chinese datasets using multilingual predictions. Although they showed that models such as GPT-4 or BERT achieve moderate to high correlations with human ratings for several but not all semantic dimensions, human ratings were based on 30-35 different responses without considering rating stability.

Thus, one methodological issue which could further impact the difference between LLM-generated and human cloze responses may lie in the stability of the measures derived from human responses. The literature provides very few indications for how many human responses are required to obtain stable, unbiased entropy estimates in cloze tasks which may impact the interpretation of experimental findings which rely on these norms to manipulate sentence constraint. For character-level predictions, Ren et al. [27] showed that ~1,000 responses per context are required for entropy convergence, validated via bootstrap analysis. To our knowledge this remains the only study to empirically quantify sample-size convergence for linguistic entropy; and, for sentence-final (word-level) tasks, no such analysis exists (Table 1). Norming studies such as Peelle et al. [2] and Angulo-Chavira et al. [28] collect ≥100 completions per sentence, reporting entropy but not evaluating stability or convergence across sample sizes. Experimental studies highly vary in the sample size for obtaining these norm estimates. While some studies rely on completions from comparable number of completions or higher (e.g., [29]: N = 72, [30]: N = 90, [31]: N = 89 – 106, [32]: N = 175, [33]: 240), other studies often use far fewer completions per sentence or do not provide any estimation (e.g., [34–41]. Similarly, Lowder & Henderson [42] and Luke & Christianson [43] aggregate large-scale word-by-word probabilities from cloze completions and eye-tracking corpora but do not include sample-size validation. Importantly, while classical work by Bormuth [44] proposes formulas for estimating sample sizes in cloze-based readability tasks based on accuracy, no guidelines exist for assessing entropy. In a recent statistical study, Arora et al. [45] suggests that entropy estimates from small samples are systematically biased.

Studies have shown that entropy or also termed predictability or sentence constraint influence not only moment-to-moment comprehension, but also higher-order processes such as semantic integration [19], syntactic ambiguity resolutions [46], and even cortical speech tracking [47]. These effects are often attenuated when model-predicted norms are used in place of human-derived values. For example, de Varda et al. [48] found stronger correlations with EEG and reading time for human cloze probabilities than for computational predictions. Similarly, Hao et al. [8] demonstrated using GPT-2 that predictability norm correlation, quantifying human-LLM alignment, outperformed perplexity in modelling behavioural data. LLM surprisal is less predictive of reading times than human cloze entropy, particularly for semantically constrained contexts [14,49]. These finding emphasize the need for empirically validated human norms.

Table 1. Sample Size Estimation and Selected Empirical Support

| Task Type | Reported N for Convergent Entropy | Comment |
| --- | --- | --- |
| Character-level prediction [27] | ~1,000 guesses/context (bootstrap convergence) | Only task with direct convergence analysis |
| Word-level sentence cloze [2,28] | ≥100 per sentence (norm; no convergence data) | Widely used norm, no validation |
| Readability (accuracy only) [44] | Formula-based (accuracy, not entropy) | Applies to proportion-correct, not entropy |
| Full-text cloze [42,43] | Large N; entropy stability not addressed | Rich probability distributions, no stopping rules |
| Estimator evaluation [18,45] | Biases identified; no N recommendation | Recommends Bayesian/bias-corrected estimators |

Taken together, there is currently no direct empirical evidence or convergence analysis available to establish the number of human responses required for reliable and unbiased entropy estimation in last-word cloze tasks. The only study in this literature set that explicitly analyses the convergence of entropy estimates as a function of sample size is Ren et al. [27], and it focuses exclusively on character-level prediction tasks. For sentence-level word prediction tasks, Peelle et al. [2] provide empirical entropy values based on ≥100 completions per sentence but offer no convergence or variability analyses. Thus, using two publicly available datasets [1,2], this study aims at answering two questions: 1) How many human responses are needed to obtain stable and unbiased entropy estimates in word-level cloze tasks (i.e., last-word sentence completions)? And 2) to what extent do entropy measures derived from LLMs align with those derived from stable human data?

2. Methods

2.1. Samples

To investigate the stability of human-based entropy for last word completions, we based our analyses on two publicly available sentence databases. The MuSe database [1] contains 619 German sentence beginnings with final-word completions from up to 232 German native speakers. The database from [2] contains 3085 English sentence beginnings with final-word completions from up to 105 participants.

Both studies applied the cloze procedure in which participants were asked to complete a sentence with the first word they expected as a sentence-final word. During the data cleaning processes, sentences with no responses and grammatical errors were excluded on a subject-level. Sentence completions were harmonized for grammatical number. For the present study, sentences were selected with at least 100 sentence completions, resulting in 476 sentences from the German and 3085 sentences from the English database, ranging from 3 to 14 words and 6 to 13 words, respectively. On average the sentence contained 158 individual responses for the German and 104 individual responses for the English database.

## 2.2. Assessing the stability of human-based entropy for last word completions
### 2.2.1. Entropy computation

To quantify the predictability of sentence completions, we computed Shannon entropy values using a bootstrap resampling approach for both, the German and English dataset. Shannon entropy [50] provides a measure of uncertainty in a probability distribution, with higher values indicating greater unpredictability in response patterns. For each sentence, we extracted all participant responses and encoded them as discrete categorical variables. Each unique response was mapped to a unique integer identifier, creating a dictionary of response categories for that sentence. For example, if participants provided responses like 'cat', 'dog', 'cat', 'bird', these would be encoded as 0, 1, 0, 2, respectively. This encoding enabled efficient computation of entropy while preserving the response distribution for each sentence.

We first computed the total entropy for each sentence across participant responses to establish an overall measure of predictability. The total entropy H was computed using the standard Shannon entropy formula (1),

(1)

$$H = -\sum_{r \in \mathcal{R}} P(r) log_2 P(r)$$

where $P(r)$ represents the empirical probability of response $r$, estimated as the observed frequency of that response divided by the total number of responses, and $\mathcal{R}$ denotes the set of total possible responses. The base-2 logarithm provides entropy values in bits, with values ranging from 0 (completely predictable; all participants gave identical responses) to $log_2 n$ where $n$ is the number of unique responses (maximum unpredictability; all responses are unique).

To assess the convergence of entropy estimates, we implemented a bootstrap simulation procedure. This approach allowed us to examine how entropy estimates evolve as more participant responses accumulate, providing insight into the minimum sample size required for stable entropy estimation. For each sentence stimulus, we conducted 1000 independent bootstrap iterations. Within each iteration, we randomly permuted the order of all participant responses and calculated cumulative entropy values as responses were added.

To improve computational efficiency, we implemented a fully vectorized permutation procedure in NumPy [51]. We generated a matrix of random values with 1000 rows corresponding to the bootstrap rounds, and as many columns as there were responses for that sentence. By sorting this matrix along the response axis and extracting the resulting indices, we obtained a matrix of permutation indices where each row corresponds to an independent random permutation of the response order. These permutation indices were then applied to reorder the encoded responses. This produced a matrix in which each row contained all responses in a different random sequence.

The integer-encoded responses were converted into one-hot representation using vectorised matrix indexing. We created a square identity matrix whose size was equal to the number of distinct response types for that sentence. Each row of this identity matrix corresponds to the one-hot encoding of one response category. By using the shuffled encoded responses as indices into this identity matrix, we obtained a three-dimensional array (number of bootstrap rounds, by total number of responses, by number of unique responses). In this representation, each response in each bootstrap round is represented by a one-hot encoded vector.

We computed cumulative sums along the response axis of the one-hot encoded array. Because the one-hot entries are binary, this operation transforms each response sequence into running counts of how often each response type has occurred. The resulting three-dimensional array maintained the same shape as the one-hot encoding matrix, with each element representing the cumulative count of a particular response up to a given response step in each bootstrap round. This provided the complete frequency distribution at every possible sample size within each bootstrap iteration.

We computed Shannon entropy in a fully vectorized manner across all bootstrap rounds and all cumulative response steps. First, count matrices were converted to probability distributions by dividing each response type's count by the total number of responses observed at that step. This normalization was performed by summing counts across all response types and dividing elementwise. Then, base-2 logarithms of the probabilities were calculated elementwise. To correctly implement Shannon's formula, undefined values arising from zero probabilities were set to 0. Entropy was then calculated across all possible responses according to formula (1). The output was an entropy matrix with 1000 rows (bootstrap rounds) and as many columns as responses, where each element represented the entropy estimate for a specific bootstrap round at a specific accumulation step.

### 2.2.2. Assessment of entropy convergence and stability

To determine the minimum sample size required to estimate sentence entropy reliably, we conducted a convergence analysis for each sentence based on the rate of change in entropy values as additional participant responses were accumulated. This analysis identified the point when entropy estimates stabilized such that collecting additional responses no longer meaningfully altered the estimated value. We defined "convergence" as the point when the rate of change (slope) of mean entropy with respect to sample size fell below 0.005 bits/response, indicating that the entropy estimate had reached a stable value.

Using the obtained entropy matrix for each sentence, we computed the mean entropy across all 1000 bootstrap iterations at each response accumulation step. This mean trajectory represented the expected entropy value at each sample size where sample size corresponds to the number of responses included so far, averaged over all re-orderings of the participant responses.

For each response step in the entropy trajectory, we calculated the slope as the change in mean entropy over a window of 5 preceding steps, expressed in units of bits per response. A step was identified as the convergence point if 1) the absolute value of the windowed slope fell below 0.005 bits/response, and 2) the slope remained below threshold for at least 3 consecutive iterations. The consecutive stability requirement ensured that the observed convergence reflected true stabilization rather than transient fluctuations. We marked as the convergence point the first step in the consecutive sequence. As a toy example, 100 sentence completions collected for a low-to-medium entropy sentence such as "She likes her tea with milk and ..." might yield three unique responses, for instance "sugar", "honey", and "stevia", with frequencies of 82, 12, and 6, respectively, corresponding to an entropy of approximately 0.85 bits. While adding a single rare and previously unobserved response such as "lemon" would increase the entropy by 0.07 bits to approximately 0.92 bits, adding another response of the most likely responses "sugar" would reduce entropy by 0.01 to approximately 0.84 resulting in a change of about 0.01 bits. As even a single rare or most likely response can lead to changes in entropy of 0.01 to 0.07 bits, which far exceeds our slope threshold of 0.005 bits/response, requiring the slope to remain below this value for several consecutive steps provides a reliable and robust definition of entropy stabilization.

## 2.3. LLM-based entropy estimation

To evaluate whether LLMs could reproduce human entropy patterns in sentence completion tasks, we estimated entropy for each sentence of the English and German databases via three different approaches: using gpt-4o via the OpenAI API [52] with 1) logit-based probability extraction, and 2) sampling-based frequency estimation, and 3) using open-source LLMs English: (gpt2-xl [53], RoBERTa Base[54]; German: german-gpt2, GottBERT [55]; both: LLaMA 2 7B Chat [56]) from Hugging Face [57]. These approaches differ in their accessibility, computational requirements, and methods for extracting probability distributions, as described below.

### 2.3.1. GPT-4o via the OpenAI API

We generated end-of-sentence completions using the gpt-4o model accessed through OpenAI API. To promote diverse completions comparable to human responses, we set the temperature parameter to 1.5, substantially higher than the default value 1.0. Higher temperature values increase the randomness of the model's output by flattening the probability distribution over possible next tokens, making rare completions more likely.

To increase comparability with human data, we prompted the model to simulate human behaviour, reflecting the natural variability observed across participants with different backgrounds and experiences. Specifically, the *system message* was the following:

"You are simulating how 100 people from different backgrounds, cultures, and ages would each intuitively complete a sentence with the *first word* that comes to their mind. Responses may vary widely in meaning, tone, connotation. Be diverse and creative."

### 2.3.2. Logit-based entropy estimation

The logit-based approach leverages the model's internal probability distribution over its vocabulary, accessing the raw logits (unnormalized log-probabilities) that the model assigns to each possible token.

We used the following two user prompts for each sentence:

*Prompt 1:* "People complete sentences differently based on their experiences, mood, and background. For the sentence: '{sentence}'. What is ONE possible word someone might use to complete it? Consider both common AND uncommon responses - the obvious answers, the creative ones, the mistakes, and the surprising choices real humans make."

*Prompt 2:* "You are simulating human sentence completions. Return EXACTLY ONE plausible next word (could be common, rare, creative, or mistaken). Rules: ONE word only, no quotes, no punctuation, no explanation. Sentence fragment: {sentence} Next word: "

With {sentence} being substituted for each sentence per generation instance.

For each sentence, we made a single API call enforcing a maximum number of 4 tokens per response., which is sufficient for most single-word completions as it accounts for multi-token words, to further guide the model to complete only with one word instead of continuing the sentence. The API response contained log-probabilities for the top 20 most likely next tokens (the maximum allowed by the API). We extracted these values and converted them to probabilities by taking the exponential of each log-probability.

Because the model operates on sub-word tokens, the same word might appear multiple times with a different tokenization (e.g.: " cat", "cat", "Cat"). To obtain word-level probabilities comparable to human responses, we normalized and aggregated the tokens, by 1) converting all tokens to lowercase, 2) stripping leading/trailing whitespaces, 3) removing all non-alphabetic characters and 4) summing probabilities for tokens that were normalized to the same word. After aggregating probabilities for identical normalized words, we renormalized the distribution to ensure it summed to 1.0. This step was necessary firstly because we only retrieved the top 20 tokens rather than the full vocabulary, and additionally because some probability mass was lost when filtering out non-alphabetic tokens.

Shannon entropy was calculated from the probability distribution using formula (1), and the SciPy library [58].

### 2.3.3. Sampling-based entropy estimation

The sampling-based approach estimates entropy by generating multiple independent completions and calculating the empirical frequency distribution of responses. This method has a higher resemblance to the human data collection procedure: just as human entropy is estimated from the observed frequency of different responses across participants, model entropy is estimated from the frequency of different completions across multiple generations. We used the same two user prompts as for the logit-based entropy estimation.

For each sentence, we generated 100 independent completions, again allowing for a maximum of 4 tokens, and additionally including as stop sequences "\n" (paragraph break) and " " (trailing space) to prevent multi-word completions. Each of the 100 generated completions was cleaned using the same normalization procedure as the logit-based approach.

We calculated the empirical frequency distribution by counting the occurrences of each unique word across the 100 completions. Shannon entropy in bits was then calculated from these empirical probabilities, as described before.

While this methodology requires 100x more generations than the logit-based approach, which increases substantially API costs, computation time and prevents reproducibility across runs, it has the advantage of being similar to human data collection and produces full words instead of tokens.

### 2.3.4. Open-source models via Hugging Face transformers library

Lastly, we evaluated several open-source language models accessible through Hugging Face's transformers library [57]. This approach offers multiple advantages, such as transparency regarding model architecture and training data and no per-query API costs. However, models that are feasible to run locally have fewer parameters and less training data than GPT-4o, potentially affecting their ability to match human response distributions.

For the English dataset, we used two models trained on English data: gpt2-xl (1.5-billion-parameter autoregressive language model) [53] and RoBERTa Base (125-million-parameter masked language model) [54]. For the German dataset, we used equivalent architectures via models trained on German data: german-gpt2 (GPT-2 autoregressive architecture) [59], and GottBERT (RoBERTa-based masked model) [55]. Additionally, we used LLaMA 2 7B Chat [56], a bigger (7 billion parameters) autoregressive language model trained on a multilingual corpus for both datasets.

All open-source models were loaded and executed using the transformers library (HuggingFace, [57]) and PyTorch [60] in Python (v3.10.16, [61]). Models were set to evaluation mode to disable dropout and other training-specific behaviours.

For autoregressive (causal) language models, we provided the input sentence fragment directly to the model, and predictions were extracted for the next token position. For masked language model, we appended a mask token to each sentence fragment to indicate the position where predictions should be generated and extracted predictions for that token.

We extracted the top 70 most likely predictions, and we normalized and aggregated tokens identically to the gpt-4o approach. After filtering and aggregation, we renormalized the probability distribution, and calculated Shannon entropy from the normalized probabilities.

### 2.3.5. Comparison metrics

To evaluate the correspondence between LLM-estimated entropy through the different methodologies and human-estimated entropy, we calculated a series of metrics for each model-dataset combination:

*Mean Absolute Error (MAE)*: Quantifying the average magnitude of discrepancies in bits, see formula (2):

(2)
$$MAE = \frac{1}{N} \sum_{i=1}^{n} |H_{model}(i) - H_{human}(i)|$$

Where $H_{model}(i)$ and $H_{human}(i)$ are the model-predicted and human-observed entropy for sentence $i$, respectively.

*Bias*: The mean signed difference between model and human entropy values, indicating systemic over- or underestimation, see formula (3):

(3)
$$Bias = \frac{1}{N} \sum_{i=1}^{n} (H_{model}(i) - H_{human}(i))$$

*Lin's Concordance Correlation Coefficient (CCC)*: Measuring agreement with the identity line (y=x), see formula (2):

(2)
$$CCC = \frac{2\rho \sigma_{human} \sigma_{model}}{\sigma_{human}^2 + \sigma_{model}^2 + (\mu_{human} - \mu_{model})^2}$$

CCC = 1.0 indicates perfect agreement (all the points fall on the identity line). Unlike Pearson's R, CCC penalizes both scatter and systematic bias or scaling differences.

To compare across German and English datasets, we subsampled randomly the English dataset to the same size as the German dataset (618 sentences).

Results

### 2.4. Stability of human-based entropy

The majority of sentences converged within the available sample size: Overall, 97.9% in the German dataset (466/476, max number of responses: 205) and 97.7% in the English dataset (3014/3085, max number of responses: 105). Both languages presented a clear positive relationship between sentence entropy and the number of sentences required for convergence: high entropy sentences require more participant responses to achieve stable entropy estimates (Fig. 1C, D). In the German dataset 90% of sentences reached convergence after 111 responses, 80% after 94 responses. Higher entropy sentences (H>2.5) needed on average 87 responses to converge, and lower entropy sentence (H<1) needed on average 19 responses to converge. Convergence points were similar in the English dataset, in which 90% of sentences reached convergence after 81 responses and 80% after 71 responses. Higher entropy sentences (H>2.5) needed on average 67 responses to converge, and lower entropy sentence (H<1) needed on average 21 responses to converge.

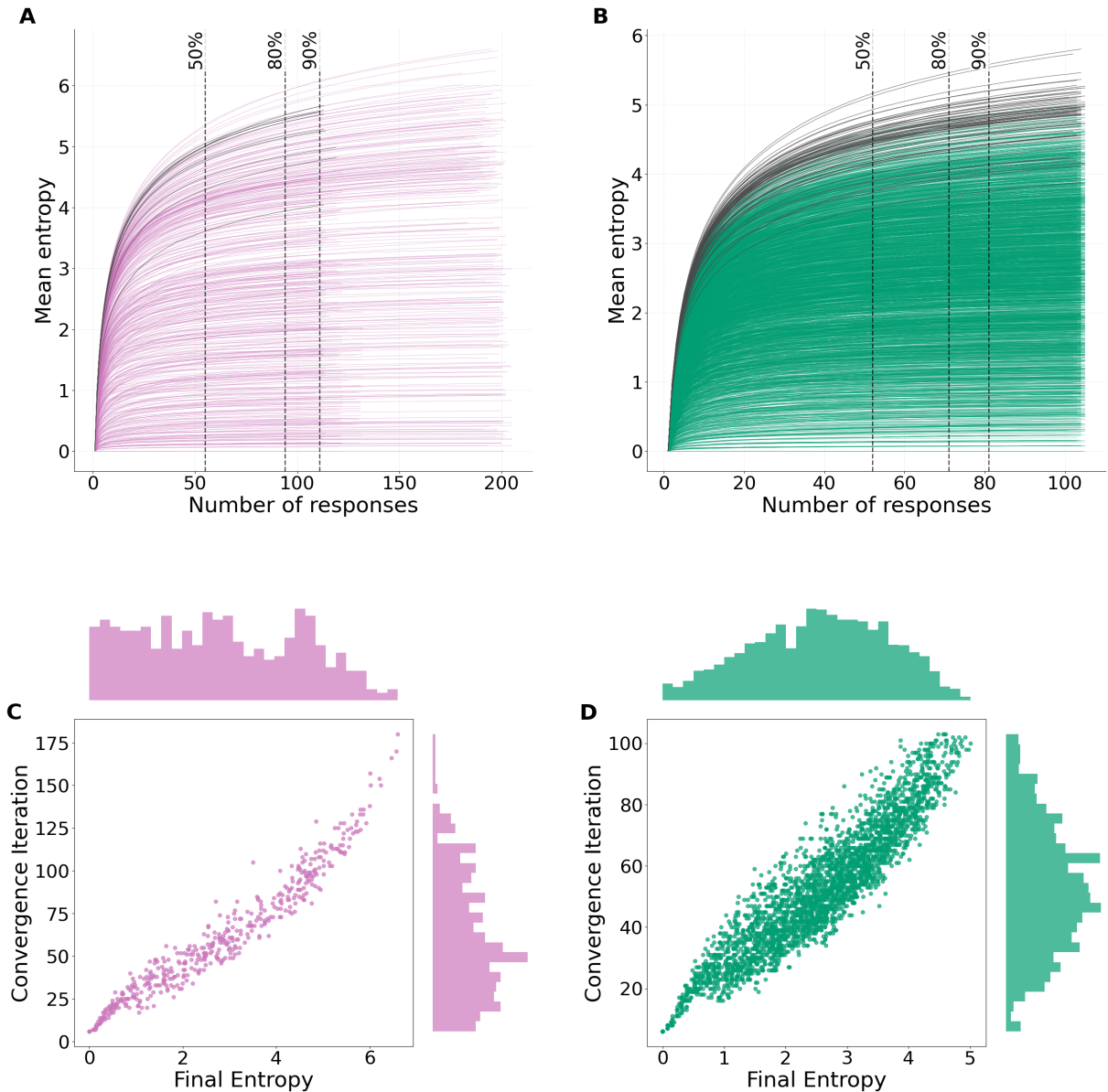

Fig 1. Entropy convergence across sentence completions. A, B) Evolution of entropy averaged across bootstrap iterations for each sentence as function of the number of collected responses in the A) German and B) English databases. In grey, sentences that did not converge. C, D) Scatter plot of the iteration at which a sentence converged against the final entropy of that sentence for the C) German and D) English database. Marginal distributions along each axis indicate the overall distributions of final entropy and convergence iteration. Only sentences that reached convergence are included. Dashed lines indicate the number of responses by which a given percentage of converged sentences reached convergence.

2.5. LLM-derived entropy and the comparison to human-based entropy

We evaluated the performance of different LLMs at predicting sentence-completion entropy in English and German datasets. Across the examined models, gpt-4o most closely resembled human predictions in both languages (Table 2). However, methodology and prompt design substantially influenced performance, particularly in the English dataset. The logit-based method achieved lower MAE (0.88 bits vs 0.98 bits), reflecting better point-wise accuracy for sentences around the centre of the entropy distribution, where the majority of sentences were concentrated. In contrast, the frequency-based method achieved higher CCC (0.513 vs 0.407), indicating better overall agreement with the human entropy values across the full range of sentence entropies (Figure 2).

Table 2. Comparison of agreement between LLM- and human- predicted entropy across different models.

| Model | Dataset used | Bias | MAE | CCC |
|---|---|---|---|---|
| gpt2-xl | English | 1.638 | 1.71 | 0.229 |
| german-gpt2 | German | 1.66 | 1.889 | 0.273 |
| roberta-base | English | 0.956 | 1.316 | 0.318 |
| GottBERT_base_best | German | 1.203 | 1.564 | 0.328 |
| Llama-2-7b-chat-hf | English | 0.4 | 1.037 | 0.467 |
| Llama-2-7b-chat-hf | German | 0.747 | 1.457 | 0.324 |
| gpt-4o logit-based (prompt 1/prompt2) | English | -0.396/-1.539 | **0.883**/1.593 | 0.407/0.227 |
| gpt-4o logit-based (prompt 1/prompt2) | German | -1.04/-1.798 | 1.3/1.833 | 0.475/0.326 |
| gpt-4o sampling-based (prompt 1/prompt2) | English | 1.064/-0.614 | 1.229/0.982 | 0.337/**0.513** |
| gpt-4o sampling-based (prompt 1/prompt2) | German | 1.407/-0.795 | 1.555/**1.052** | 0.424/**0.689** |

*Note: MAE: Mean Absolute Error; CCC: Lin's Concordance Correlation Coefficient. In bold, the minimum achieved MAE and CCC for both English and German datasets.*

We illustrate the differences between gpt-4o and human sentence completion by examining sentences with the largest entropy discrepancies, both positive and negative, using the frequency-based method (Fig. 2A, lower). The largest positive error (+3.03 bits) corresponded to the sentence: "The mother was against swearing, calling it a __". While humans were highly uniform in their responses, completing with "sin" in 86.4% of the cases (entropy = 1.03 bits), gpt-4o generated a wide range of completions, with its most frequent response appearing in only 23% of samples (4.07 bits). The largest negative error (-3.99 bits) occurred for the sentence: "The elite country club was luxurious and __". Here, humans were more diverse in their responses, with "expensive" appearing in only 26.0 % of responses (4.29 bits), while gpt-4o showed high certainty, generating "exclusive" in 96% of samples (0.30 bits).

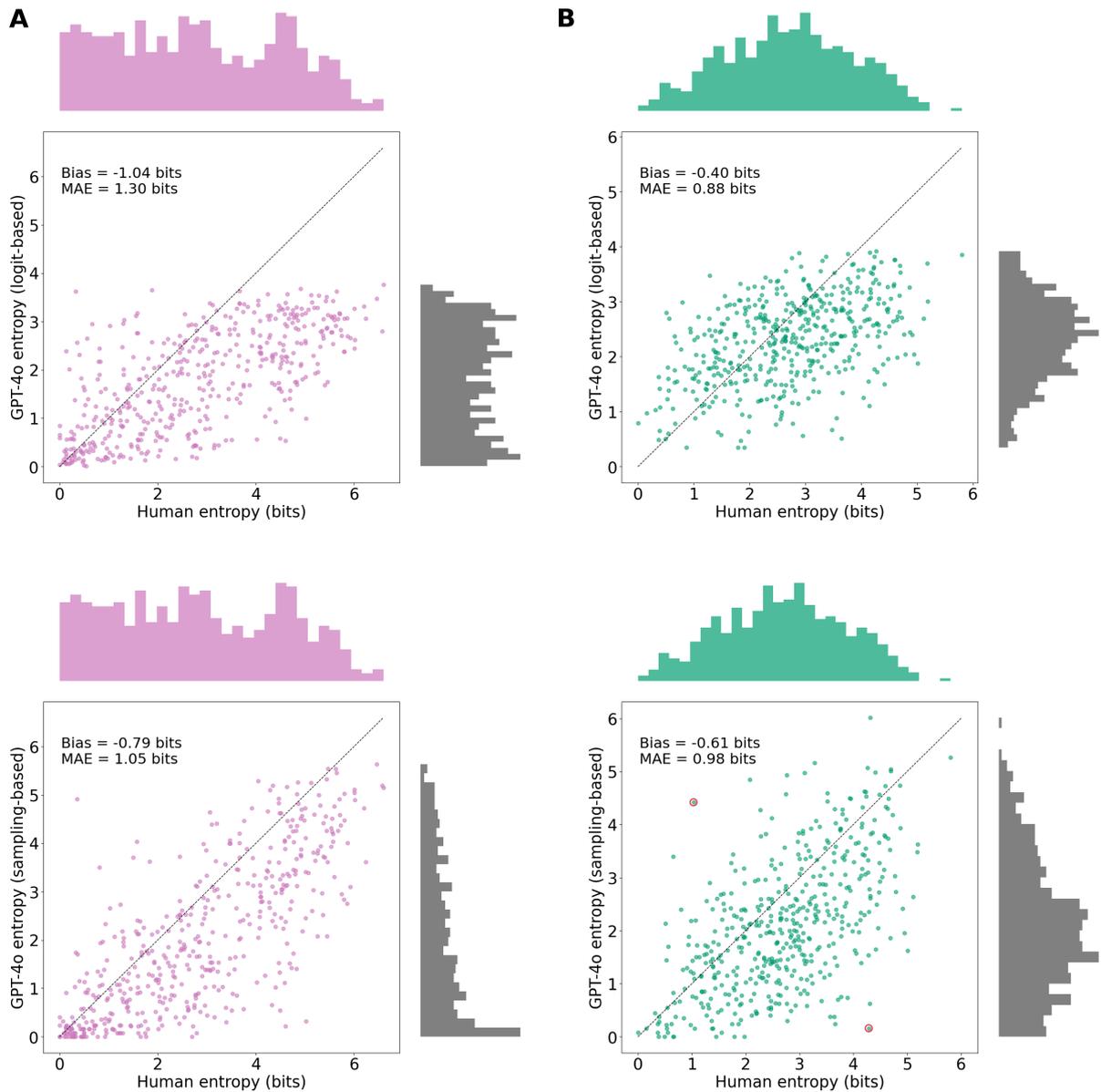

**Fig 2. Comparison of human- and model predicted sentence completion entropy.** Scatter plots the best gpt-4o- against human-predicted entropy for A) German and B) English datasets. Upper plots reflect the logit-based method (prompt 1) and lower the sampling-based (prompt 2). The identity line indicates perfect agreement. Red circled points mark sentences with the largest positive (gpt-4o entropy > human entropy) and negative (human entropy > gpt-4o entropy) error. Marginal distributions along each axis indicate the overall distributions of both predictions. Notice the high density of 0 bits (maximum certainty) predictions by the LLM. English dataset was subsampled to match the German dataset in size.

3.  Discussion

This study addressed two central methodological and theoretical questions in psycholinguistics: (1) How many human responses are required to obtain stable, reliable estimates of sentence-level entropy in cloze tasks? And (2), to what extent can entropy values derived from large language models (LLMs) approximate these human-derived benchmarks? By analysing large-scale German and English datasets with rigorous convergence analyses and systematic LLM comparisons, we provide empirical guidance for both research practice and the interpretation of computational models.

Our first major finding is that more than 97% of entropy estimates derived from human cloze responses stabilise after collecting 205 responses for the complete German dataset and 105 responses for the complete English dataset, while 90% convergence is already achieved after just 111 and 81 responses, respectively. While this provides a first direct empirical justification for the previously untested convention in norming studies (e.g. [2,28]), it also emphasises that these measures are language specific, making norming studies necessary across different languages. Furthermore, convergence is not uniform: High-entropy (semantically less predictable) sentences systematically required more responses to stabilise than low-entropy (semantically predictable) ones, confirming theoretical expectations about sampling bias [45,62]. This pattern resolves an apparent conflict with character-level studies, which suggest a need for orders of magnitude more samples [27]. The discrepancy is theoretically meaningful: word-level sentence completions are sharply constrained by semantic, syntactic, and pragmatic context, leading to faster convergence than in character-level or unconstrained prediction tasks. This finding has direct implications for norming practices. While collecting ~100 responses per sentence for an English dataset and ~120 for a German dataset is generally sufficient for aggregate analysis, a fixed-N approach systematically under-samples the most unpredictable contexts. Future studies could therefore adopt adaptive sampling strategies, continuing data collection until an entropy-based stability criterion is met, thereby increasing efficiency and statistical rigor.

Our second core finding is that LLM-derived entropy shows a substantial, though imperfect, correspondence with stable human estimates. Among tested models, GPT-4o achieved the strongest alignment in both languages. This result aligns with work showing relationships between computational predictability measures and human behaviour [8,17]. Crucially, however, the degree of alignment depended on the method of entropy extraction and the design of the prompt. We show that small changes in the prompt lead to vast differences and that the effectiveness of a prompt also depends on further analysis strategies. Logit-based estimation from the model's probability distribution minimized point-wise error (MAE), favouring numerical accuracy near the mean. In contrast, entropy estimated by sampling multiple model completions yielded higher concordance (CCC), better capturing the full scale and dispersion of human variability across sentences. This suggests that simulating response variability makes the model's output distribution more comparable to the inherently more diverse human distribution.

Despite promising aggregate correlations, persistent qualitative discrepancies have been reported. Many studies show that LLMs frequently displayed over-dispersion in contexts where humans strongly agreed on a dominant completion, and over-confidence where human responses were diverse. Our estimations show over-confidence in LLMs compared to human ratings. This systematic misalignment, in which LLMs underweight the dominant human response and overweight atypical alternatives confirms previous descriptions (e.g., [14,18,19]). It indicates that while LLMs may approximate how constrained a context is (entropy), they often differ on which specific completions are plausible and the shape of the probability distribution; or are unable to mimic highly individual, experience-based, and potentially creative completions. These distributional differences carry significant theoretical weight. Entropy alone can mask profound structural differences: two distributions can share similar entropy while differing markedly in skew, modality, or tail weight [62]. Our findings thus highlight the necessity of moving beyond scalar entropy to richer distributional comparisons, such as divergence measures (e.g., Kullback-Leibler divergence) when comparing human and model representations.

The root of these discrepancies likely lies in the foundational differences in how semantics are acquired. Human semantic knowledge emerges from embodied, multimodal, and socially grounded experience [63,64]. LLMs, in contrast, derive their representations exclusively from statistical patterns in text [65]. While this process captures a remarkable amount of linguistic constraint, it lacks the experiential component. Consequently, even when entropy values roughly align, the underlying probability structures reflect fundamentally different representational principles. This is consistent with evidence that LLM semantic spaces are less richly interconnected and more homogenized than human semantic networks [23,24].

Our results lead to practical guidelines: For human norming, a minimum of 100 responses per sentence is generally sufficient for stable word-level entropy, but researchers should be cautious of high-entropy items and consider adaptive sampling, keeping in mind potential language-related differences. For using LLMs as proxies, particularly for the strongest models like GPT-4o, LLMs can serve as valuable approximations for human entropy, but they are not drop-in replacements. The choice of model and extraction method must be strategically aligned with the research goal: While logit-based entropy (e.g., from GPT-4o) should be used when the priority is point-wise numerical accuracy , such as when entropy is used as a continuous predictor in regression modelling, sampling-based entropy should be used when the goal is to capture the human response distributions, as it better represents the variability, dispersion, and distributional shape. In contrast to stronger GPT-4 models, smaller or less capable models (e.g., GPT-2, RoBERTa) show significantly poorer agreement and should be used with caution.

Several limitations should be considered for future research: (1) while 100 responses sufficed for most sentences in our datasets, extremely open-ended semantic contexts may require larger samples. (2) our focus was on sentence-final word prediction; entropy dynamics for ongoing, incremental prediction or phrase-level completions may differ. (3) LLM performance is tied to specific inbuild structures and design of prompts; systematic exploration of these parameters could refine alignment.

Future work should therefore (1) implement and test adaptive human data collection protocols; (2) compare degrees of LLM-human entropy alignment to predictive power for behavioural and neural data (e.g., reading times, EEG); and (3) explore whether prompting strategies or parameter tuning can make LLM distributions more human-like.

## Conclusion

In summary, this study provides the first direct evidence that human sentence-level entropy stabilises with approximately 100 cloze responses, validating a common practice while highlighting its limits for high-entropy items. Furthermore, we demonstrate that modern LLMs can approximate human entropy to a meaningful degree, but this alignment is highly method-dependent and coexists with systematic qualitative differences in distributional structure. LLMs are thus powerful tools for modelling contextual predictability, but their application requires careful methodological choices and theoretical scrutiny. Human data remain the gold standard for capturing the nuanced, experientially grounded structure of semantic expectations.